\documentclass[runningheads]{llncs}

\usepackage{graphicx}
\usepackage{booktabs}
\usepackage{multirow}
\usepackage{subcaption}
\usepackage{amsmath}
\usepackage{hyperref}
\usepackage{url}
\usepackage{xcolor}
\usepackage{siunitx}
\usepackage{enumitem}
\usepackage{microtype}

\begin{document}

\title{A Semi-Automated Framework for 3D Reconstruction of Medieval Manuscript Miniatures}

\author{Riccardo Pallotto\inst{1}\orcidID{0009-0007-4422-3328} \and
Pierluigi Feliciati\inst{1}\orcidID{0000-0002-2499-8528} \and
Tiberio Uricchio\inst{2}\orcidID{0000-0003-1025-4541}}

\authorrunning{R. Pallotto et al.}

\institute{University of Macerata, Macerata, Italy\\
\email{\{r.pallotto, pierluigi.feliciati\}@unimc.it}
\and
University of Pisa, Pisa, Italy\\
\email{tiberio.uricchio@unipi.it}}

\maketitle

\begin{abstract}
This paper presents a semi-automated framework for transforming two-dimensional miniatures from medieval manuscripts into three-dimensional digital models suitable for extended reality (XR), tactile 3D~printing, and web-based visualization. We evaluate seven image-to-3D methods (TripoSR, SF3D, SPAR3D, TRELLIS, Wonder3D, SAM~3D, Hi3DGen) on 69~manuscript figures from two collections using rendering-based metrics (Silhouette IoU, LPIPS, CLIP~Score) and volumetric measures (Depth Range Ratio, watertight percentage), revealing a trade-off between volumetric expansion and geometric fidelity. Hi3DGen~\cite{hi3dgen2025} balances topological quality with rich surface detail through its normal bridging approach, making it a good starting point for expert refinement. Our pipeline combines SAM~\cite{kirillov2023segment} segmentation, Hi3DGen mesh generation, expert refinement in ZBrush, and AI-assisted texturing. Two case studies on Gothic illuminations from the \textit{Decretum Gratiani} (Vatican Library) and Renaissance miniatures by Giulio Clovio demonstrate applicability across artistic traditions. The resulting models can support WebXR visualization, AR overlay on physical manuscripts, and tactile 3D~prints for visually impaired users.

\keywords{Medieval manuscripts \and 3D reconstruction \and image-to-3D generation \and quantitative evaluation \and extended reality \and digital cultural heritage}
\end{abstract}

\section{Introduction}

The digitization of cultural heritage has transformed access to historical artifacts~\cite{deegan2002digital}, yet digital representations are often perceived as mere surrogates~\cite{robertson2020digital}. We argue for a different perspective: digital manuscripts should function as \emph{complements} that enrich their physical counterparts~\cite{paul2020curation,campagnolo2020book}. Lindh{\'e}~\cite{lindhe2015materiality} calls this \emph{digital materiality}: the idea that digital objects can activate imagination and creativity through embodied interaction. When combined with 3D modeling and extended reality (XR) technologies, manuscript illuminations can transcend their two-dimensional origins to become interactive objects that can be experienced through WebXR, augmented reality overlays, virtual exhibitions, and tactile 3D prints for visually impaired users~\cite{haynes2023iiif3d}.

This paper presents a semi-automated framework for generating 3D digital models from illuminated manuscript miniatures. Our approach leverages foundation models for segmentation~\cite{kirillov2023segment} and neural 3D reconstruction~\cite{hi3dgen2025}, while maintaining the human-in-the-loop expertise essential for historical fidelity. The resulting models are what Baker~\cite{baker2025paradata} calls \emph{memory twins}: digital representations that preserve cultural memory while enabling novel interactions. The framework itself is not the end goal but rather an operational tool designed to accelerate the production of refined 3D models, which remain the central output of this work and still require expert intervention to ensure formal quality and art-historical coherence.

Our contributions are: (1)~a \emph{quantitative evaluation} of seven image-to-3D methods on 69 medieval manuscript figures using rendering-based (Silhouette IoU, LPIPS, CLIP) and volumetric metrics; (2)~a fully specified \emph{pipeline} with reproducible automated stages and documented manual intervention, combining SAM segmentation, Hi3DGen mesh generation, expert refinement, and AI-assisted texturing; (3)~two \emph{case studies} on Gothic and Renaissance manuscripts with XR-ready outputs.

\section{Related Work}

\subsection{Digital Manuscripts and 3D Cultural Heritage}

Three-dimensional digitization of cultural heritage typically employs photogrammetry~\cite{remondino2011heritage} and structured light scanning~\cite{pavlidis2007methods,pietroni2023innovative,pietroni2023codex4d,brown2001acquisition}, which require physical 3D form. For two-dimensional sources such as manuscript illuminations, different approaches are needed. Recent work has explored depth estimation from single images using neural networks~\cite{ke2024marigold}, as well as feed-forward 3D reconstruction from single images~\cite{hong2023lrm}. However, digitized manuscripts are not mere surrogates but ``raw digital data to be organized, analyzed, annotated, aggregated, explored, linked, manipulated, and visualized''~\cite{paul2020curation}. Endres~\cite{endres2019digitizing} demonstrates through his 3D reconstruction of the St.~Chad Gospels that digital objects are rhetorical objects that actively participate in knowledge construction. Our work contributes to this perspective by introducing 3D as a new modality.

\subsection{Image-to-3D Generation Methods}

Single-image 3D reconstruction has advanced rapidly. TripoSR~\cite{tochilkin2024triposr} predicts triplane NeRF representations; SF3D~\cite{boss2024sf3d} and SPAR3D~\cite{huang2025spar3d} improve upon this with UV-unwrapping and point-aware reconstruction, respectively; TRELLIS~\cite{xiang2025trellis} uses structured 3D latent diffusion; Wonder3D~\cite{long2024wonder3d} fuses multi-view normals via NeuS~\cite{wang2021neus}; SAM~3D~\cite{liang2025sam3d} combines segmentation-aware features with 3D generation. Hi3DGen~\cite{hi3dgen2025} uses \emph{normal bridging}, estimating surface normals as intermediate 2.5D representations to guide geometry generation, preserving fine details such as facial features and drapery folds. A common limitation is that these methods are trained on photographic imagery; their behavior on stylized medieval miniatures has not been thoroughly evaluated.

\section{Methodology}

\begin{figure}[t]
  \centering
  \includegraphics[width=\linewidth]{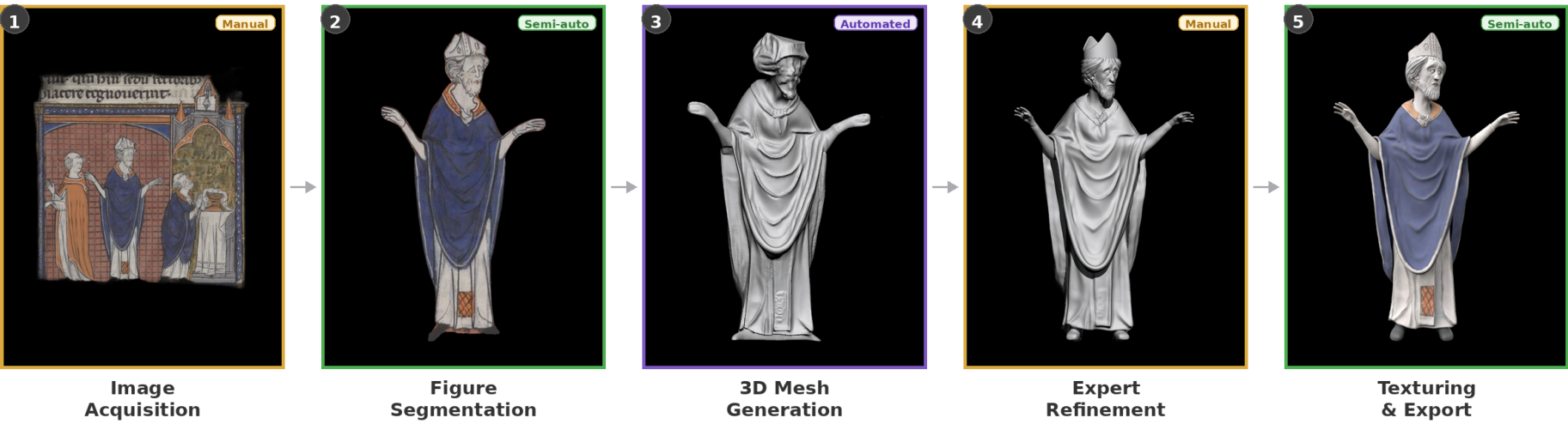}
  \caption{Pipeline overview: image acquisition, SAM segmentation, Hi3DGen mesh generation, expert refinement, and AI-assisted texturing.}
  \label{fig:pipeline_overview}
\end{figure}

Our framework (Fig.~\ref{fig:pipeline_overview}) combines automated AI processing with expert manual intervention, reflecting the principle that cultural heritage applications require both computational efficiency and scholarly accuracy. The pipeline consists of five stages: image acquisition, segmentation, 3D mesh generation, expert refinement, and texturing/export. Following Baker~\cite{baker2025paradata} and the concept of digital storytelling~\cite{bonacini2020musei}, we create \emph{imaginative reconstructions} that bring illustrated figures into spatial form.

\paragraph{Segmentation.} Individual figures are isolated from manuscript illuminations using SAM~\cite{kirillov2023segment} with the ViT-H backbone (632M parameters) and point prompts. For each figure, we provide three foreground point prompts placed at the head, torso, and base of the figure. When these do not capture fine details such as extended hands or attributes (e.g., a bishop's crosier), we add 1--2 additional points. The segmented elements are exported as RGBA PNG images with transparent backgrounds.

\paragraph{3D Mesh Generation.} Segmented figures are processed through Hi3DGen~\cite{hi3dgen2025}, which operates in two stages: sparse structure generation (guidance strength 7.5, 50 sampling steps) producing a coarse voxel grid, followed by structured latent generation that refines it into a dense mesh. Input resolution is $1024 \times 1024$~px; source images are automatically downscaled. Generation requires approximately 12~GB VRAM.

\paragraph{Expert Refinement.} The raw meshes produced by Hi3DGen, while geometrically plausible, require expert refinement to correct artifacts and ensure historical appropriateness. Using ZBrush, we perform: (i)~\emph{artifact correction} of anatomical errors, merged geometry, and missing details; (ii)~\emph{retopology} from high-polygon output (typically 200k--800k faces) to clean mesh structures suitable for real-time rendering (15k--60k faces); (iii)~\emph{proportion adjustment} to preserve medieval stylistic conventions that differ from naturalistic representation; and (iv)~\emph{detail enhancement} of facial features, hands, and textile patterns. This human-in-the-loop step is essential: while AI can generate plausible 3D forms, ensuring that these forms are appropriate representations of medieval art requires art-historical expertise. Refinement time decreases with operator experience as subsequent figures from the same manuscript share stylistic conventions.

\paragraph{Texturing and Export.} After retopology, the models undergo UV unwrapping. Source textures are upscaled $4\times$ (Real-ESRGAN~\cite{wang2021realesrgan}), expanded generatively to fill occluded regions (e.g., the back of a cope), and manually refined in Substance Painter\footnote{\url{https://www.adobe.com/products/substance3d/apps/painter.html}}. The model is rendered with an unlit shader to preserve the original painterly appearance. Models are exported as GLB for WebXR, OBJ for archiving, and FBX for game engines.

\section{Experimental Evaluation}

This section evaluates the automated 3D generation stage, i.e., the output of the seven methods \emph{before} any manual refinement. The effect of the full pipeline, including expert refinement, is assessed in Section~\ref{sec:case_studies}.

\subsection{Dataset and Metrics}

We evaluate on two test sets: 38 segmented figures from Monteprandone manuscripts (Civic Museum of Monteprandone, various codices from the library of San~Giacomo della~Marca) and 31 from the \textit{Decretum Gratiani} (Cod.~Ross.~Lat.~308, Vatican Library, 14th~century), for a total of 69~figures. Figures are segmented using SAM as described in Section~3. Seven methods are compared with default parameters.

Since no ground-truth 3D models exist for medieval manuscript miniatures, the only available reference is the original 2D image. We therefore combine rendering-based metrics that compare front-view renders against the input with volumetric measures that characterize geometric properties directly:

\paragraph{Orientation detection.} Each method produces meshes in different coordinate systems. We automatically detect the viewing direction by identifying the thinnest bounding box axis, then test 16~candidate in-plane orientations (2~directions $\times$ 4~rotations $\times$ 2~flips), selecting the one maximizing CLIP similarity with the reference among candidates with IoU above 50\% of the maximum.

\paragraph{Rendering-based 2D metrics.} Since not all methods produce textured outputs, we standardize comparison by rendering each mesh without texture using a uniform diffuse material under fixed lighting from the automatically detected front view. We compute: \emph{Silhouette IoU} (overlap between reference alpha mask and rendered depth mask); LPIPS~\cite{zhang2018lpips} (learned perceptual distance, AlexNet features); and CLIP Score~\cite{radford2021clip} (semantic consistency, ViT-B/32 embeddings), both computed on gray-background composites. Traditional pixel-level metrics (PSNR, SSIM) are ill-suited here: the systematic color mismatch between untextured gray renders and colorful painted miniatures dominates both metrics, preventing meaningful discrimination between methods. All rendering-based metrics are computed from the automatically detected front view, which is the natural perspective for figures derived from 2D illuminations; the volumetric metrics complement this by characterizing full 3D geometry independently of viewpoint.

\paragraph{Volumetric 3D metrics.} These quantify geometric properties without requiring ground-truth 3D: \emph{Depth Range Ratio}, defined as $\text{depth} / \max(\text{width}, \text{height})$ of the axis-aligned bounding box (values near 0 indicate flat, relief-like geometry; values approaching 1 indicate fully three-dimensional objects); and \emph{watertight mesh percentage}, the fraction of outputs forming closed, manifold surfaces suitable for 3D printing and robust real-time rendering.

\subsection{Results}

\begin{table*}[t]
  \caption{Quantitative comparison on two manuscript datasets. Best in \textbf{bold}, second-best \underline{underlined}. Depth Range Ratio should be read jointly with watertight percentage (see Section~\ref{sec:depth_analysis}).}
  \label{tab:comparison}
  \centering
  \footnotesize
  \begin{tabular}{ll@{\hspace{6pt}}c@{\hspace{6pt}}c@{\hspace{6pt}}c@{\hspace{6pt}}c@{\hspace{6pt}}c}
    \toprule
    & \textbf{Method} & \textbf{IoU}$\uparrow$ & \textbf{LPIPS}$\downarrow$ & \textbf{CLIP}$\uparrow$ & \textbf{Depth R.} & \textbf{WT\%}$\uparrow$ \\
    \midrule
    \multirow{7}{*}{\rotatebox[origin=c]{90}{\small\textit{Monteprandone}}}
    & TripoSR~\cite{tochilkin2024triposr}   & 0.459 & 0.547 & 0.721 & \underline{0.371} & 0\% \\
    & SF3D~\cite{boss2024sf3d}               & \textbf{0.751} & \textbf{0.395} & 0.724 & 0.234 & 0\% \\
    & SPAR3D~\cite{huang2025spar3d}          & \underline{0.734} & \underline{0.399} & 0.725 & 0.262 & 0\% \\
    & TRELLIS~\cite{xiang2025trellis}        & 0.572 & 0.457 & 0.716 & 0.041 & 39\% \\
    & Wonder3D~\cite{long2024wonder3d}       & 0.348 & 0.522 & 0.677 & \textbf{0.513} & 16\% \\
    & SAM~3D~\cite{liang2025sam3d}           & 0.594 & 0.437 & \underline{0.730} & 0.048 & \textbf{68\%} \\
    & Hi3DGen~\cite{hi3dgen2025}             & 0.557 & 0.431 & \textbf{0.744} & 0.190 & \underline{53\%} \\
    \midrule
    \multirow{7}{*}{\rotatebox[origin=c]{90}{\small\textit{Vatican}}}
    & TripoSR~\cite{tochilkin2024triposr}   & 0.795 & \underline{0.589} & \underline{0.592} & \underline{0.653} & 0\% \\
    & SF3D~\cite{boss2024sf3d}               & \underline{0.832} & 0.660 & 0.576 & 0.134 & 0\% \\
    & SPAR3D~\cite{huang2025spar3d}          & \textbf{0.846} & 0.658 & 0.549 & 0.135 & 0\% \\
    & TRELLIS~\cite{xiang2025trellis}        & 0.790 & 0.658 & 0.558 & 0.115 & 13\% \\
    & Wonder3D~\cite{long2024wonder3d}       & 0.665 & 0.676 & 0.521 & \textbf{0.777} & 6\% \\
    & SAM~3D~\cite{liang2025sam3d}           & 0.773 & 0.675 & 0.569 & 0.064 & \textbf{71\%} \\
    & Hi3DGen~\cite{hi3dgen2025}             & 0.704 & \textbf{0.577} & \textbf{0.689} & 0.406 & \underline{68\%} \\
    \bottomrule
  \end{tabular}
\end{table*}

Table~\ref{tab:comparison} reveals consistent patterns across both datasets. SF3D and SPAR3D achieve the highest silhouette overlap (IoU 0.751/0.734 on Monteprandone, 0.832/0.846 on Vatican), indicating that their feed-forward architectures best match the source silhouette. Hi3DGen achieves the highest CLIP scores on both datasets (0.744, 0.689), suggesting stronger semantic preservation, and the best LPIPS on Vatican (0.577). Wonder3D achieves the highest depth ratios (0.513, 0.777), indicating volumetric geometry, but with poor topology (16\%, 6\% watertight) and the lowest IoU on Monteprandone (0.348). SAM~3D achieves the highest watertight rate (68--71\%) but produces flat geometry (depth ratios 0.048--0.064). Hi3DGen balances geometric soundness with moderate volumetric expansion (0.190--0.406) and preserves fine surface detail (see Section~\ref{sec:depth_analysis}).

\subsection{Geometric Fidelity vs.\ Volumetric Expansion}
\label{sec:depth_analysis}

The Depth Range Ratio reveals that a higher depth ratio does not indicate better reconstruction. Wonder3D produces the highest depth ratios (0.513, 0.777) but predominantly non-watertight meshes and the lowest silhouette IoU on Monteprandone (0.348); TripoSR follows (0.371, 0.653) with zero watertight output. Conversely, SAM~3D, TRELLIS, and Hi3DGen produce topologically sound meshes (39--71\% watertight) but with conservative depth. On the Vatican dataset, Hi3DGen stands out with a depth ratio of 0.406 while maintaining 68\% watertight output.

We attribute these patterns to architectural differences: Wonder3D's NeuS-based SDF reconstruction produces volumetric but topologically inconsistent geometry; TripoSR's triplane NeRF hallucinates back geometry aggressively; SAM~3D and TRELLIS use structured latent diffusion, producing sound topology but conservative depth. Hi3DGen's normal bridging approach preserves essential surface articulation (drapery folds, facial features, body volumes) that other methods smooth away. For expert refinement, richly articulated surfaces can be scaled outward and refined far more efficiently than high-depth but featureless geometry. To validate this assessment, we complement the automated metrics with a perceptual study.

\subsection{Perceptual Evaluation (User Study)}
\label{sec:user_study}

We conducted a Two-Alternative Forced Choice (2AFC) study: in each trial, a participant viewed the original 2D miniature alongside geometry-only renders from two methods and selected the one that better matched the original in shape and detail similarity. All pairwise comparisons were generated for each evaluation figure. Twenty-six volunteers completed 3{,}686 trials, ensuring full pair coverage.

\begin{table}[t]
  \caption{2AFC perceptual evaluation (3{,}686 trials, 26 participants). Win\% = fraction of trials in which the method was preferred.}
  \label{tab:user_study}
  \centering
  \footnotesize
  \begin{tabular}{ll@{\hspace{8pt}}r@{\hspace{4pt}}c@{\hspace{4pt}}r@{\hspace{8pt}}r}
    \toprule
    & \textbf{Method} & \textbf{Wins} & / & \textbf{Total} & \textbf{Win\%}$\uparrow$ \\
    \midrule
    \multirow{7}{*}{\rotatebox[origin=c]{90}{\small\textit{Monteprandone}}}
    & Hi3DGen~\cite{hi3dgen2025}             & 489 & / & 585 & \textbf{83.6} \\
    & TripoSR~\cite{tochilkin2024triposr}   & 403 & / & 569 & \underline{70.8} \\
    & SAM~3D~\cite{liang2025sam3d}           & 324 & / & 569 & 56.9 \\
    & SF3D~\cite{boss2024sf3d}               & 267 & / & 564 & 47.3 \\
    & SPAR3D~\cite{huang2025spar3d}          & 263 & / & 577 & 45.6 \\
    & TRELLIS~\cite{xiang2025trellis}        & 215 & / & 562 & 38.3 \\
    & Wonder3D~\cite{long2024wonder3d}       &  27 & / & 550 & 4.9 \\
    \midrule
    \multirow{7}{*}{\rotatebox[origin=c]{90}{\small\textit{Vatican}}}
    & Hi3DGen~\cite{hi3dgen2025}             & 417 & / & 492 & \textbf{84.8} \\
    & TripoSR~\cite{tochilkin2024triposr}   & 394 & / & 484 & \underline{81.4} \\
    & SAM~3D~\cite{liang2025sam3d}           & 303 & / & 483 & 62.7 \\
    & TRELLIS~\cite{xiang2025trellis}        & 249 & / & 490 & 50.8 \\
    & SPAR3D~\cite{huang2025spar3d}          & 155 & / & 481 & 32.2 \\
    & SF3D~\cite{boss2024sf3d}               & 157 & / & 490 & 32.0 \\
    & Wonder3D~\cite{long2024wonder3d}       &  23 & / & 476 & 4.8 \\
    \bottomrule
  \end{tabular}
\end{table}

Hi3DGen achieves the highest preference on both datasets (83.6\%, 84.8\%), followed by TripoSR (70.8\%, 81.4\%). Wonder3D is consistently last ($<$5\%). Inter-rater agreement was found moderate to strong (Kendall's $W = 0.70$, $p < 0.001$). These rankings corroborate the automated metrics and provide independent human validation for Hi3DGen as the preferred starting point for expert refinement.

\section{Case Studies}
\label{sec:case_studies}

\begin{figure*}[t]
  \centering
  \includegraphics[width=\textwidth]{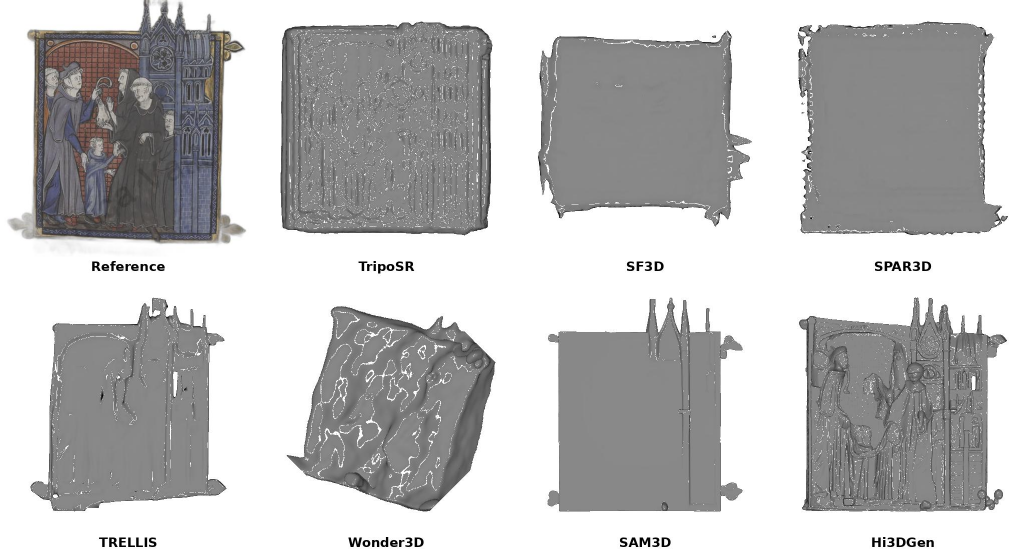}
  \caption{Qualitative comparison on a Vatican figure (bishop from \textit{Decretum Gratiani}). Hi3DGen best preserves surface detail; Wonder3D produces the most volumetric but topologically flawed geometry; SAM~3D produces clean watertight meshes but almost flat.}
  \label{fig:comparison_vat}
\end{figure*}

\begin{figure*}[t]
  \centering
  \includegraphics[width=\textwidth]{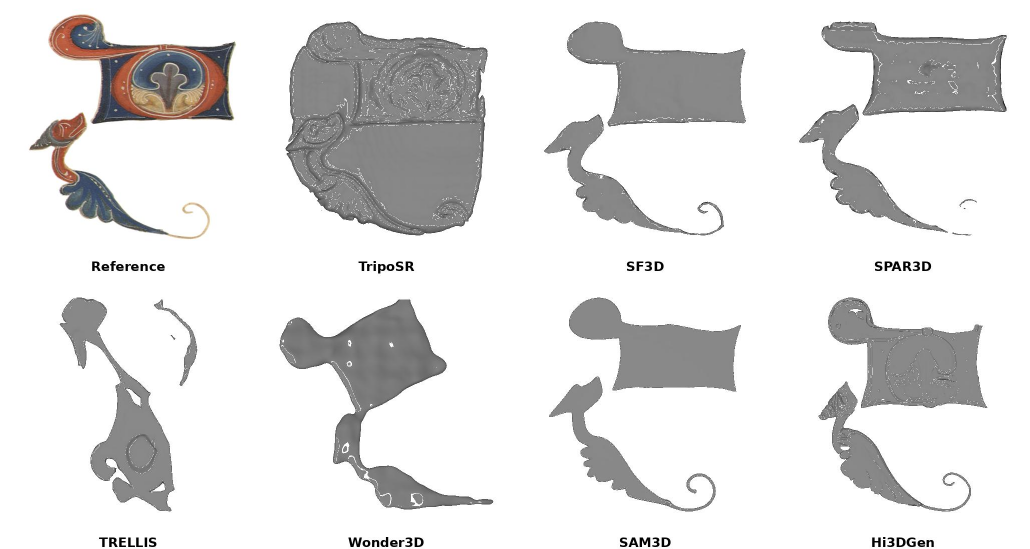}
  \caption{Qualitative comparison on a Monteprandone figure. SF3D and SPAR3D produce the closest silhouette match; Hi3DGen preserves surface articulation and semantic content.}
  \label{fig:comparison_mont}
\end{figure*}

\begin{figure}[t]
  \centering
  \begin{subfigure}[t]{0.55\linewidth}
    \centering
    \includegraphics[width=\linewidth]{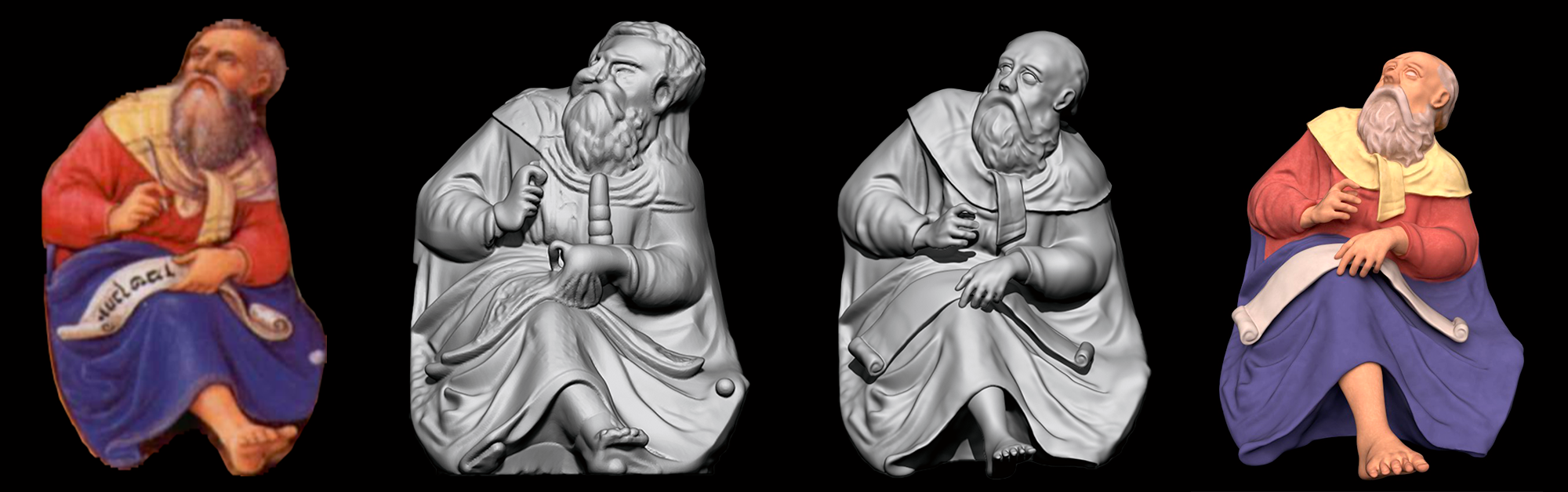}
    \caption{Prophet from the Colonna Missal.}
    \label{fig:clovio}
  \end{subfigure}
  \hfill
  \begin{subfigure}[t]{0.41\linewidth}
    \centering
    \includegraphics[width=\linewidth]{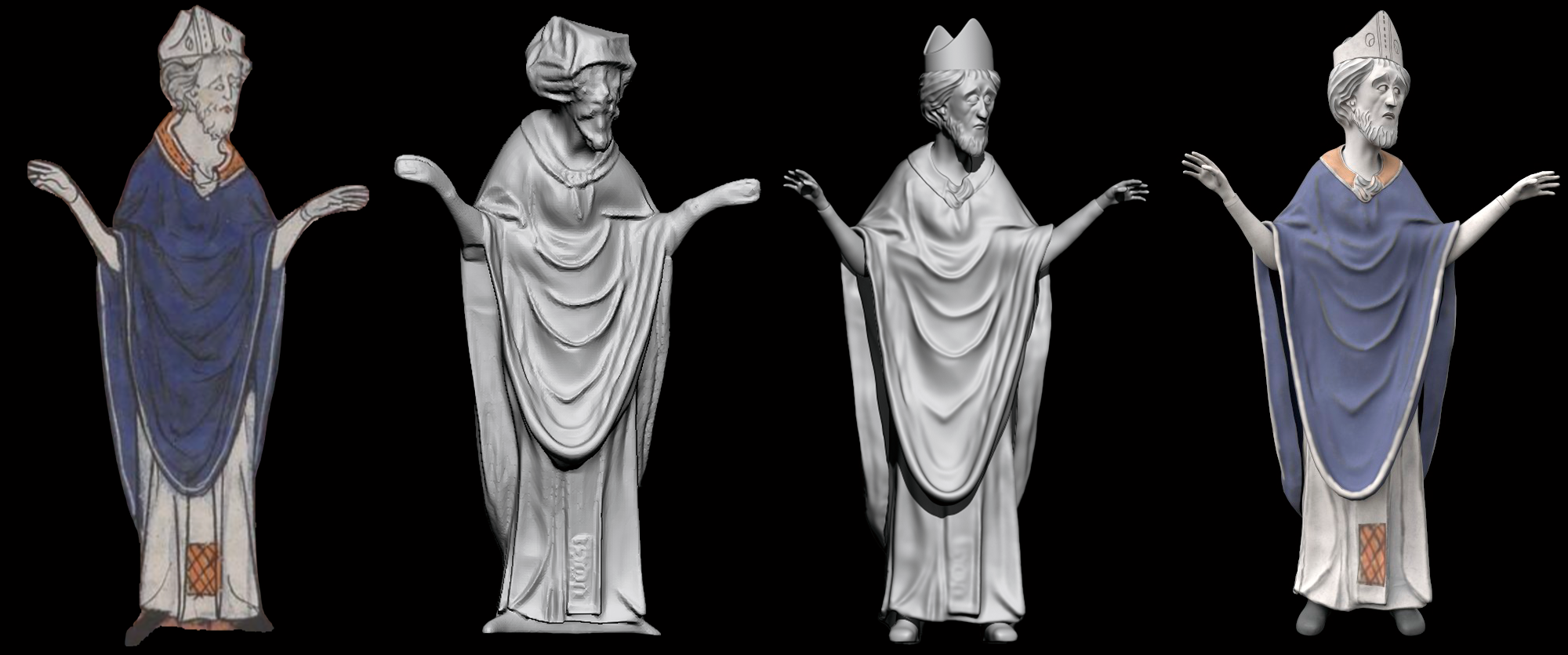}
    \caption{Bishop from the \textit{Decretum Gratiani}.}
    \label{fig:bishop}
  \end{subfigure}
  \caption{Pipeline stages: original miniature, Hi3DGen output, refined mesh, textured model.}
  \label{fig:results}
\end{figure}

We demonstrate the complete pipeline, from segmentation through expert refinement and texturing, on two manuscripts representing different artistic traditions (Fig.~\ref{fig:comparison_vat},~\ref{fig:comparison_mont},~\ref{fig:results}).

\paragraph{Gothic: The Decretum Gratiani.} The \textit{Decretum Gratiani} (Cod.~Ross.~Lat.~308) is a 14th-century manuscript held by the Vatican Library, originating from the library of San Giacomo della Marca in Monteprandone~\cite{siliquini2002decorazione}. The original collection comprised 700--800 volumes; today, 61 remain in the Civic Museum of Monteprandone, while others are dispersed across institutions including the Vatican Library. This dispersal makes digital reunification through 3D models particularly meaningful for scholars, the local community of Monteprandone, and a broader public interested in medieval heritage. We applied the full pipeline to a bishop figure from a decorated initial: SAM segmentation with three point prompts isolated the figure from the ornamental border; Hi3DGen produced an initial mesh of $\sim$184k faces; sculpting refinement in ZBrush corrected anatomical artifacts and reduced the mesh to $\sim$158k faces, further retopologized to $\sim$30k polygons for real-time rendering while preserving the Gothic drapery linearity. Finally, textures were projected from the high-resolution source image and refined in Substance Painter.

\paragraph{Renaissance: The Colonna Missal.} To test generalizability across artistic styles, we applied the same pipeline to miniatures by Giulio Clovio (1498--1578)~\cite{cionini1980clovio} from the Colonna Missal (John Rylands Library). Clovio's miniatures present different challenges: naturalistic anatomy, complex foreshortening, and sophisticated chiaroscuro modeling. After SAM segmentation, Hi3DGen produced an initial mesh of $\sim$831k faces, retopologized in ZBrush to $\sim$36k polygons. Hi3DGen performed more consistently on these Renaissance figures, likely because the naturalistic proportions align more closely with AI training data. Manual refinement was still necessary for hands and stylistic details.

\begin{table}[t]
  \caption{Per-figure time breakdown for the semi-automated pipeline.}
  \label{tab:time_breakdown}
  \centering
  \footnotesize
  \begin{tabular}{lrl}
    \toprule
    \textbf{Pipeline Step} & \textbf{Time} & \textbf{Type} \\
    \midrule
    Image acquisition         & $\sim$5 min  & Manual \\
    SAM segmentation          & $\sim$2 min  & Automated \\
    Hi3DGen mesh generation   & $\sim$3 min  & Automated \\
    Manual refinement (ZBrush)& $\sim$12 h   & Manual \\
    Texturing \& export       & $\sim$2 h    & Semi-automated \\
    \midrule
    \textbf{Total (semi-automated)} & $\sim$\textbf{14 h} & --- \\
    Fully manual baseline     & $\sim$27 h   & Manual \\
    \bottomrule
  \end{tabular}
\end{table}

\paragraph{Effect of manual refinement.} The primary benefits of expert refinement are qualitative and structural: artifact correction (anatomically implausible hands, merged geometry), historical appropriateness (drapery linearity, iconographic details), watertight topology suitable for 3D printing, and polygon optimization for real-time rendering (from $\sim$184k to $\sim$30k faces for the bishop, $\sim$831k to $\sim$36k for the prophet).

\paragraph{Applications.} The GLB exports enable WebXR browser-based viewing without plugins, AR overlay on physical manuscript pages, and virtual museum integration. Both models have also been 3D-printed as tactile reproductions for visually impaired users~\cite{francomano2022access}.

\paragraph{Scalability.} Table~\ref{tab:time_breakdown} details the per-step time breakdown. The primary bottleneck is manual refinement ($\sim$12~h/figure), while the automated stages (segmentation + generation) require only $\sim$5 minutes per figure and scale linearly. Based on our two case studies, the semi-automated pipeline ($\sim$14~h/figure) reduces production time by $\sim$48\% compared to fully manual modeling ($\sim$27~h, estimated by having the same operator model both figures entirely from scratch without any AI-generated starting mesh). Refinement time decreases with operator experience, and advances in AI-assisted sculpting may further reduce manual effort.

\section{Conclusion}
\vspace{-1em}
Through quantitative evaluation of seven image-to-3D methods on 69 manuscript figures, we identified trade-offs between volumetric expansion and geometric fidelity. SF3D and SPAR3D achieve the best silhouette overlap, Hi3DGen the highest semantic fidelity, while Hi3DGen balances topological quality with rich surface detail. A 2AFC perceptual study confirmed Hi3DGen as the preferred method, with moderate to strong inter-rater agreement, providing independent human validation for the choice of starting point for expert refinement. The semi-automated pipeline reduces production time by $\sim$48\% compared to fully manual modeling, with outputs suitable for WebXR, AR, and tactile accessibility. 

\begin{credits}
\subsubsection{\ackname}
Work partially supported by the Italian Ministry of Education and Research (MUR) in the framework of the FoReLab project (Departments of Excellence). We acknowledge NVIDIA Corporation for the DGX Spark system donated through the NVIDIA Academic Grant Program.
\end{credits}
\vspace{-1em}
\bibliographystyle{splncs04}
\bibliography{references}

\end{document}